\documentclass[
  pre,
  twocolumn,
  twoside,
  byrevtex,
  superscriptaddress,
  floatfix,
  longbibliography,
  nofootinbib
]{revtex4-2}

\usepackage[realmainfile]{currfile}
\usepackage{amssymb}
\usepackage{amsmath}
\usepackage{listings}
\usepackage{xcolor}

\definecolor{electricblue}{rgb}{0.0, 0.4, 1.0}
\definecolor{brightred}{rgb}{1.0, 0.0, 0.0}
\definecolor{gray}{rgb}{0.4, 0.4, 0.4}

\lstdefinestyle{cleanpython}{
  language=Python,
  basicstyle=\fontsize{10}{9.5}\selectfont\ttfamily,  
  keywordstyle=\color{electricblue}\bfseries,
  commentstyle=\color{gray}\itshape,
  stringstyle=\color{brightred},
  showstringspaces=false,
  numbers=none,
  frame=none,
  backgroundcolor=\color{white},
  breaklines=true,
  tabsize=4
}

\usepackage{url}
\PassOptionsToPackage{hyphens}{url}\usepackage{hyperref}
\makeatletter
\g@addto@macro{\UrlBreaks}{\UrlOrds}
\makeatother

\definecolor{goodblue}{RGB}{0, 91, 187}
\hypersetup{
  colorlinks=true,
  allcolors=goodblue,
  urlcolor=goodblue,
  citecolor=goodblue,
  pdfborder={0 0 0},
  breaklinks=true,
}

\usepackage[normalem]{ulem}

\makeatletter

\def\CT@@do@color{%
  \global\let\CT@do@color\relax
  \@tempdima\wd\z@
  \advance\@tempdima\@tempdimb
  \advance\@tempdima\@tempdimc
  \advance\@tempdimb\tabcolsep
  \advance\@tempdimc\tabcolsep
  \advance\@tempdima2\tabcolsep
  \kern-\@tempdimb
  \leaders\vrule
  \hskip\@tempdima\@plus  1fill
  \kern-\@tempdimc
  \hskip-\wd\z@ \@plus -1fill }
\makeatother

\definecolor{olivegreen}{rgb}{0.33333,.41961,0.18431}
\definecolor{forestgreen}{rgb}{0.13333,.5451,0.13333}
\definecolor{lightgrey}{rgb}{0.7,0.7,0.7}
\definecolor{verylightgrey}{rgb}{0.90,0.90,0.90}
\definecolor{veryverylightgrey}{rgb}{0.95,0.95,0.95}
\definecolor{grey}{rgb}{0.5,0.5,0.5}

\definecolor{headerblue}{HTML}{33367E}
\definecolor{unitednationsblue}{HTML}{4D88FF}

\definecolor{charcoal}{HTML}{36454F}
\definecolor{cinerous}{HTML}{98817B}
\definecolor{feldgrau}{HTML}{4D5D53}
\definecolor{glaucous}{HTML}{6082B6}
\definecolor{arsenic}{HTML}{3B444B}
\definecolor{xanadu}{HTML}{738678}

\definecolor{firebrick}{HTML}{B22222}
\definecolor{orangered}{HTML}{FF4500}
\definecolor{tomato}{HTML}{FF6347}

\definecolor{purpletaupe}{HTML}{3B444B}

\newcommand{\done}[1]{}

\newenvironment{textblock}{\renewcommand{\item}{}\ignorespaces}{}

\usepackage{titlesec}
\titleformat*{\paragraph}{\bfseries}

\usepackage{graphicx}
\usepackage{epsfig}
\usepackage{verbatim}

\usepackage{enumerate}
\usepackage{enumitem}

\usepackage{amssymb}
\usepackage{amsmath}

\usepackage{ifthen}

\usepackage{longtable}

\usepackage{mathtools}

\newboolean{twocolswitch}

\usepackage{array}

\newcommand{\PreserveBackslash}[1]{\let\temp=\\#1\let\\=\temp}

\newcommand{\sindex}[1]{}
\newcommand{\nindex}[1]{}

\newcommand{\www}[1]{\url{#1}}

\usepackage{lettrine}

\newcommand{\lmat}{\left[
    \begin{array}
    }
    \newcommand{\rmat}{\end{array}
  \right]
}

\newcommand{\sizerank}{r}

\newcommand{\bigrank}{R}

\newcommand{\indexaraw}{1}
\newcommand{\indexbraw}{2}

\newcommand{\elementsymbol}{\tau}

\newcommand{\rtd}[1]{D^{\textnormal{R}}_{#1}}
\newcommand{\rtdelement}[1]{\delta D^{\textnormal{R}}_{#1,\elementsymbol}}

\newcommand{\rtdalpha}{\rtd{\alpha}}

\newcommand{\rtdnorm}{\mathcal{N}_{\indexaraw,\indexbraw;\alpha}}
\newcommand{\invrtdnorm}{\frac{1}{\rtdnorm}}

\newcommand{\bigrankordering}{\bigrank_{\indexaraw,\indexbraw;\alpha}}

\begin{document}
\setboolean{twocolswitch}{true}



\raggedright

\title{\protect
  A suite of allotaxonometric tools
for the comparison of complex systems
using rank-turbulence divergence


}

\author{
  \firstname{Jonathan}
  \surname{St-Onge}
}
\email{These authors contributed equally.}

\affiliation{
  Vermont Complex Systems Institute,
  University of Vermont,
  Burlington,
  VT 05405,
  USA.
}

\affiliation{
  Joint lab, 
  University of Vermont,
  Burlington, 
  VT 05405, 
  USA.
}

\author{
  \firstname{Ashley M. A.}
  \surname{Fehr}
}
\email{These authors contributed equally.}

\affiliation{
  Vermont Complex Systems Institute,
  University of Vermont,
  Burlington,
  VT 05405,
  USA.
}

\affiliation{
  Computational Story Lab,
  University of Vermont,
  Burlington,
  VT 05405,
  USA.
}

\author{
  \firstname{Carter}
  \surname{Ward}
}

\affiliation{
  Vermont Complex Systems Institute,
  University of Vermont,
  Burlington,
  VT 05405,
  USA.
}

\affiliation{
  Computational Ethics Lab,
  University of Vermont,
  Burlington,
  VT 05405,
  USA.
}

\author{
  \firstname{Calla G.}
  \surname{Beauregard}
}

\affiliation{
  Vermont Complex Systems Institute,
  University of Vermont,
  Burlington,
  VT 05405,
  USA.
}

\affiliation{
  Computational Story Lab,
  University of Vermont,
  Burlington,
  VT 05405,
  USA.
}

\author{
  \firstname{Michael V.}
  \surname{Arnold}
}

\affiliation{
  Vermont Complex Systems Institute,
  University of Vermont,
  Burlington,
  VT 05405,
  USA.
}

\affiliation{
  Computational Story Lab,
  University of Vermont,
  Burlington,
  VT 05405,
  USA.
}

\author{
  \firstname{Samuel F.}
  \surname{Rosenblatt}
}

\affiliation{
  Vermont Complex Systems Institute, 
  University of Vermont,
  Burlington, 
  VT 05405, 
  USA.
}

\affiliation{
  Joint lab, 
  University of Vermont,
  Burlington, 
  VT 05405, 
  USA.
}

\author{
  \firstname{Benjamin}
  \surname{Cooley}
}

\affiliation{
  Vermont Complex Systems Institute,
  University of Vermont,
  Burlington,
  VT 05405,
  USA.
}

\author{
  \firstname{Christopher M.}
  \surname{Danforth}
}

\affiliation{
  Vermont Complex Systems Institute,
  University of Vermont,
  Burlington,
  VT 05405,
  USA.
}

\affiliation{
  Computational Story Lab,
  University of Vermont,
  Burlington,
  VT 05405,
  USA.
}

\affiliation{
  Department of Mathematics \& Statistics,
  University of Vermont,
  Burlington,
  VT 05405,
  USA.
}

\author{
  \firstname{Peter Sheridan}
  \surname{Dodds}
}
\email{peter.dodds@uvm.edu}

\affiliation{
  Vermont Complex Systems Institute,
  University of Vermont,
  Burlington,
  VT 05405,
  USA.
}

\affiliation{
  Computational Story Lab,
  University of Vermont,
  Burlington,
  VT 05405,
  USA.
}

\affiliation{
  Department of Computer Science,
  University of Vermont,
  Burlington,
  VT 05405,
  USA.
}

\affiliation{
  Santa Fe Institute,
  1399 Hyde Park Rd,
  Santa Fe,
  NM 87501,
  USA.
}

\date{\today}

\begin{abstract}
  \protect
  \begin{textblock}
\item
  Describing and comparing complex systems requires principled,
  theoretically grounded tools. 
\item
  Built around the phenomenon of type turbulence,
  allotaxonographs provide map-and-list visual comparisons
  of pairs of heavy-tailed distributions.
\item
  Allotaxonographs are designed to accommodate
  a wide range of instruments
  including
  rank- and probability-turbulence divergences,
  Jenson-Shannon divergence,
  and
  generalized entropy divergences.
\item
  Here, we describe a suite of programmatic tools
  for rendering
  allotaxonographs
  for rank-turbulence divergence
  in 
  Matlab, 
  Javascript, 
  and 
  Python, 
  all of which have different use cases.
\end{textblock}
 
\end{abstract}

\maketitle




\setlength{\parskip}{1\baselineskip plus .1\baselineskip  minus .1\baselineskip}

\titlespacing*{\section}{0pt}{\baselineskip}{\baselineskip}
\titlespacing*{\subsection}{0pt}{0.5\baselineskip}{0.5\baselineskip}

\section{Introduction}
\label{sec:papertag.introduction}

\begin{textblock}
\item
  Complex systems are often comprised of types
  which vary in `size' over many orders of magnitude~\cite{zipf1949a,newman2005power,coromina-murtra2010a}.
\item
  Size is usually a number
  such as
  counts of word types in a corpus~\cite{zipf1949a},
  populations of cities,
  species abundance in an ecosystem,
  baby names in a culture,
  citations to scientific papers~\cite{redner1998_citations},
  and
  total wealth of individuals or corporations.
\item
  Studying such systems requires tools that can contend with
  both massive size variation
  and type turbulence.
\item
  Type turbulence refers to how
  component rankings broadly vary 
  between adjacent complex systems~\cite{pechenick2017a,dodds_allotaxonometry_2023}.
\item
  Rankings tend to differ little for the highest ranked components
  (e.g., the most common words in two English language books).
\item
  But as we move to increasingly lower ranked components,
  we typically see more and more `turbulence' in rankings,
  with the flux of components across rank thresholds
  scaling superlinearly with rank~\cite{pechenick2017a}.
\item
  Rank-turbulence divergence (RTD)
  is a visualization instrument designed
  for the comparison of any two ranked lists of
  heavy-tailed distributions
  which exhibit type turbulence.
\item
  We define RTD below and refer the reader to 
  the foundational paper,
  Ref.~\cite{dodds_allotaxonometry_2023},
  for the full conceptual and analytic development of the RTD instrument
  along with allotaxonographs.
\item 
  For RTD, we only need two rankings over a set of types.
\item
  Though these rankings will be derived from
  some underlying type sizes,
  we may not have access to them
  (e.g., Amazon ranks items by sales
  but does not release sales numbers).
\end{textblock}

\begin{figure}[tp!]
  \centering	
  \includegraphics[width=0.82\columnwidth]{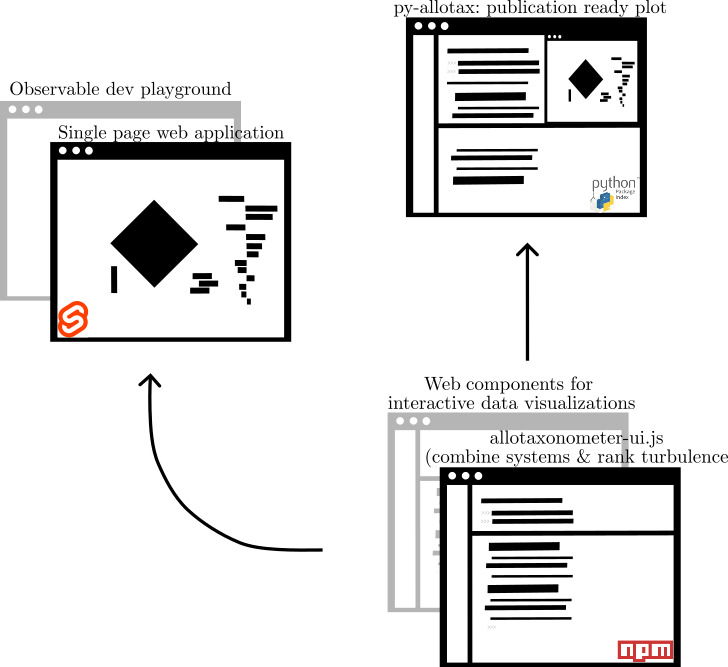}
  \caption{
    The suite of Allotaxonometer tools.
    Not included here is the original Matlab tool~\cite{matlab_allotax}.
  }
  \label{fig:papertag.ecosystem}
\end{figure}

\begin{textblock}
\item
  In this brief paper,
  we report on our efforts to increase access to the study of allotaxonometry
  by providing a suite of complementary
  avenues of approach for rank-turbulence divergence (RTD).
\item
  First, we describe the original Matlab tool
  that, amongst other things,
  handles the largest amounts of data~\cite{matlab_allotax}.
\item
  Second, we provide a single-page web application that allows users to experiment with RTD allotaxonographs,
  a visual representation of the allotaxonometry instrument: 
  \href{https://complex-stories.uvm.edu/allotaxonometry}{complex-stories.uvm.edu/allotaxonometry}~\cite{webapp_allotax_2}.

\item 
  Third, we introduce \texttt{allotaxonometer-ui.js},
  a JavaScript package~\cite{js_allotax}.
\item
  Finally, for the Python community,
  we offer the \texttt{py-allotax} tool~\cite{py_allotax}. 
\item 
  This ecosystem of tools,
  which we lay out in
  Fig.~\ref{fig:papertag.ecosystem},
  affords varying levels of accessibility.
\end{textblock}

\begin{textblock}
\item
  The definition of rank-turbulence divergence is as follows.
\item
  We denote the rank of component $\elementsymbol$
  in system $i$ by $\sizerank_{\elementsymbol,i}$.
\item
  When working from sizes, we use tied ranking.
\item
  All components that do not appear in one system but do appear in the other
  will be given a tied last rank.
\item
  Rank-turbulence divergence has a single tunable parameter,
  $0 \le \alpha < \infty$,
  which allows the user to adjust
  for the level of type turbulence present between two systems.
\item
  As $\alpha$ is tuned from 0 to $\infty$,
  the instrument will go from surfacing rare types
  to the most common types as the strongest contributors
  to the overall divergence score.
\item
  The definition of RTD is:
\item
  \begin{align}
    &
    \rtdalpha(
    \bigrank_{\indexaraw}
    \,\|\,
    \bigrank_{\indexbraw}
    )
    =
    \sum_{\elementsymbol \in \bigrankordering}
    \rtdelement{\alpha}(
    \bigrank_{\indexaraw}
    \,\|\,
    \bigrank_{\indexbraw}
    )
    \nonumber
    \\
    &
    =
    \invrtdnorm
    \frac{\alpha+1}{\alpha}
    \sum_{\elementsymbol \in \bigrankordering}
    \left\lvert
    \frac{1}{\left[\sizerank_{\elementsymbol,\indexaraw}\right]^{\alpha}}
    -
    \frac{1}{\left[\sizerank_{\elementsymbol,\indexbraw}\right]^{\alpha}}
    \right\rvert^{1/(\alpha+1)}.
    \label{eq:allotaxtools.rankturbdiv}
  \end{align}
\item 
  The normalization factor 
  $\rtdnorm$ is determined by taking
  the two rankings to be from systems
  with no shared types,
  resulting in 
\item 
  $
  0
  \le
  \rtdalpha(
  \bigrank_{\indexaraw}
  \,\|\,
  \bigrank_{\indexbraw}
  )
  \le
  1.
  $
\end{textblock}
 
\begin{figure*}[tp!]
  \centering	
  \includegraphics[width=0.95\textwidth]{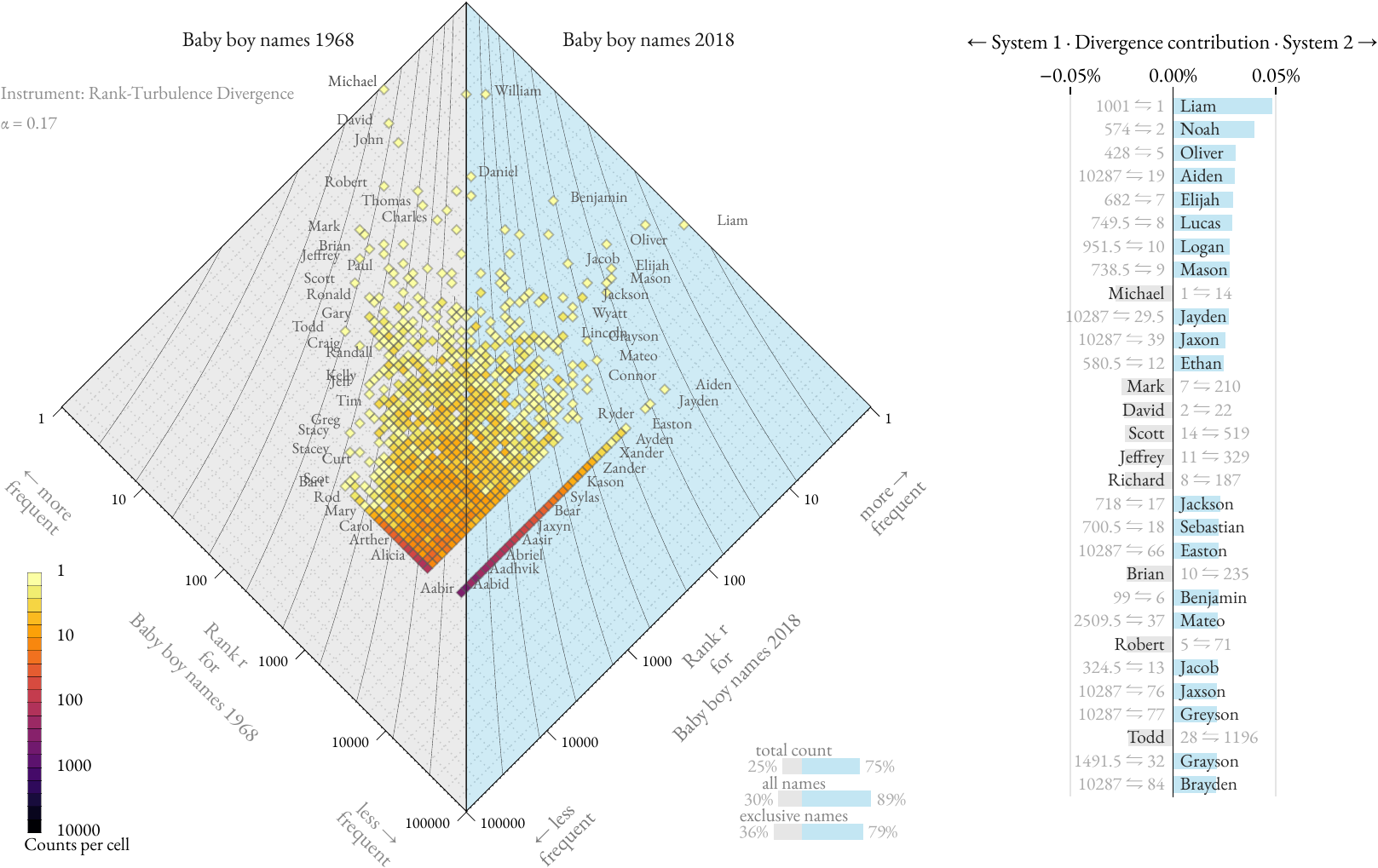}  
  \caption{
    The centerpiece of the allotaxonograph, the diamond plot,
    with a colormap (bottom left), balance plot (bottom right), and wordshift plot (right).
    Allotaxonographs compare
    two complex systems through their heavy-tailed size rankings of types while attending
    to the distribution scaling of the systems' counts.
    Allotaxonographs are composed of a map-like histogram in logarithmic rank space
    and a list of the types most strongly contributing to
    the overall divergence.
    The particular instrument we use here is rank-turbulence divergence.
    See Sec.~2.2 in Ref.~\cite{dodds_allotaxonometry_2023} 
    for details on how to read all aspects of allotaxonographs.
  }
  \label{fig:papertag.diamond}
\end{figure*}

\begin{textblock}
\item
  In this paper, we describe this suite of tools for studying allotaxonometry, outlining uses, and accessibility.
\item
  Inspired by work in visual story telling
  and designing primarily for user experience (UX), 
\item
  we propose that accessibility is both
\item
  a quality of the code 
\item
  (for instance, following good software practices) 
\item
  and phenomenological 
\item
  (letting people directly experience complex ideas
  such as heavy-tailed distributions).
\item
  We conclude with future work and lessons learned from broadening the allotaxonometry toolset 
\item
  through the diversification of our computational toolbox.
\end{textblock}

\section{Tools}
\label{sec:papertag.tools}

\subsection{Matlab}

\begin{textblock}
\item
  The first allotaxonometer toolkit runs on Matlab~\cite{dodds_allotaxonometry_2023}.
\item
  Used with Matlab running on a local system,
  this tool enables the comparison of much larger scale datasets than the tools described next
  (10s and 100s of millions of types).
\item
  The Matlab tool accommodates zero probabilities 
  so that probability-based divergences can be meaningfully visualized,
\item 
  which is a feature not yet incorporated into 
  rank-based allotaxonographs we introduce here for different platforms.
\item
  Users can choose from a range of instruments
  including
  rank-turbulence divergence,
  probability-turbulence divergence~\cite{dodds2020g},
  Jensen-Shannon divergence~\cite{lin1991a},
  and
  generalized entropy divergences~\cite{cichocki2010a}.
\item
  Probability-turbulence divergence itself is a general tool
  that includes a number of independent divergences as special cases:
  The S{\o}rensen-Dice coefficient~\cite{dice1945a,sorensen1948a,looman1960a},
  the $F_{1}$ score~\cite{vanrijsbergen1979a,sasaki2007a};
  the Hellinger distance~\cite{hellinger1909a};
  the Mautusita distance~\cite{matusita1955a};
  the Motyka distance~\cite{deza2006a};
  and 
  all $L^{(p)}$-norm type constructions.
\item 
  We gather online resources for the Matlab version,
  including papers and links to its Gitlab repository~\cite{matlab_allotax}.
\end{textblock}

\subsection{JavaScript package}

\begin{textblock}
\item
  The main allotaxonometer package is a JavaScript
  package~\cite{js_allotax} designed to facilitate the reuse of
  allotaxonograph calculations and charting components.
\item 
  This package's calculations are used as input to the web tool and
  the \texttt{py-allotax} tool.
\item
  In their own coding environments, users can use
  \texttt{npm} (a package manager) to install:
\item 
  \texttt{npm install allotaxonometer-ui}.
\end{textblock}

\begin{textblock}
\item
  To experiment with changing the \texttt{allotaxonometer}
  calculation or graphing methods, we recommend the Observable
  framework's notebooks\footnote{\url{https://observablehq.com/platform/notebooks}.} as offering components for
  creating interactive tutorials\footnote{An Observable notebook demonstrating the Allotaxonometer: \url{https://observablehq.com/@jstonge/allotaxonometer-4-all}.}, supporting
  the interactive and educational dissemination of allotaxonometry.
\item
  Observable notebooks are similar in style and functionality to
  Jupyter or Python notebooks which allow cell execution, notes, and
  visualization.
\end{textblock}

\subsection{Web application}

\begin{textblock}
\item
  The Allotaxonometer web tool~\cite{webapp_allotax_2} works directly in browser, allowing users to immediately enter data, explore, and compare systems. 
\item
  The tool uses \texttt{allotaxonometer-ui.js} calculations and implements reusable web components for interactive storytelling.
\end{textblock}

\begin{textblock}
\item
  Users follow the link and land on a single page in which they upload two data files, one for each system of comparison.
\item 
  Users can upload many files and toggle which systems to compare while adjusting the alpha ($\alpha$) value parameter for calculating rank-turbulence divergence. 
\item
  Lastly, users can click to download a static image of the displayed system. 
\end{textblock}

\subsection{Python package}

\begin{textblock}
\item
  The Python package \texttt{py-allotax}~\cite{py_allotax} 
  enables a programmatic approach to allotaxonometry for researchers
  and programmers to generate graphs in their own coding environments.
\item 
  The tool supports queries of data up to about 2 GB.
\end{textblock}

\begin{textblock}
\item
From Github or PyPi, users follow installation instructions for JavaScript and Python dependencies. 
\item
  Users can convert data to the required format or provide data and arguments to generate the HTML and/or PDFs of the allotaxonograph. 
\item
  Static image generation is done from the command line, a script, or a Python notebook.
\item
  The code example below compares US baby boy names\footnote{\href{https://catalog.data.gov/dataset/baby-names-from-social-security-card-applications-national-data}{Baby names from social security card applications national data (2023) from the US Social Security Administration.}} recorded in 1968 (system 1) and 2018 (system 2) 
  by providing:
  \texttt{json} files,
\item
  a save location for the graph, an alpha value, and 2 system titles
  to display, respectively.
\end{textblock}
\begin{lstlisting}
from py_allotax.generate_svg import generate_svg

generate_svg(
    "boys_1968.json",
    "boys_2018.json",
    "output_location/notebook_test.pdf",
    "0.17",
    "Baby boy names 1968",
    "Baby boy names 2018"
)
\end{lstlisting}
\begin{textblock}
\item
  In the repository, users can find additional information on using
  Python virtual environments,
\item 
  an example notebook, and frequently asked questions.
\end{textblock}
\section{Concluding remarks}
\label{sec:papertag.concludingremarks}
\begin{textblock}
\item
  The study of allotaxonometry and how systems
  compare has wide-ranging research use cases~\cite{dodds_allotaxonometry_2023}. 
\item
  This project aimed to enhance allotaxonometry study while ensuring
  our tools stand the test of time.
\item
  Our goal was to enable users to create engaging web stories
  while providing assurance that core
  functionalities will remain reproducible.
\item
  What have we learned in the process? What guarantees can we make to our users?
\end{textblock}





\begin{textblock}
\item
    Upon designing \texttt{py-allotax}, we faced a `two language problem'
  in deciding how to prioritize user accessibility while maintaining
  project maintainability and distributed software components.
\item 
    By two language problem, we mean reconciling tooling from the interactive world with scientific programming while allowing custom web components.
\item 
    We leverage the Svelte framework (\url{https://svelte.dev/}) that allows writing hybrid code that is both server-side renderable (enabling a static version callable from Python) and served on the client (taking advantage of modern browser environments). 
\item
    By providing all ecosystem components (calculations and visualization), the \texttt{allotaxonometer-ui} is more testable, extendable, and changes distribute downstream. 
\item 
    Yet, we recognize that this scientific work is typically done in programming languages such as Python which host large ecosystems for handling matrix operations.
\item
  Ultimately we learned the software components could not be
  modular as desired without sacrificing ease of installation
  and access for researchers.
\item
  Accessibility remained our primary goal to ensure researchers can
  get started using these tools immediately with a shallow learning curve for
  installation.
\end{textblock}


\vspace{-1em}
\begin{textblock}
\item
  A next step
  would be the accommodation of the many divergences available for probability distributions.
\item 
  The main feature needed is for allotaxonographs to have a separate
  section along both histograms' bottom axes for 
  zero probabilities ($-\infty$ in logarithmic space~\cite{dodds2020g,dodds_PTD_flipbooks_2025}).
\item 
  In doing so, some care is required to connect contour lines from the main body of the histogram to the separate section.
\item
  Other welcome contributions involve
  reducing JavaScript or Python ecosystem dependencies and optimizing the performance of producing static images from HTML.
\end{textblock}


\acknowledgments

\begin{textblock}
\item
  Our work was supported in part by
  National Science Foundation Award \#242829
  (Science of Online Corpora, Knowledge, and Stories),
 \item 
 the Alfred P. Sloan Foundation (G-2024-22498),
\item 
  a foundational gift from MassMutual,
\item 
  and a philanthropic gift from an anonymous source.
\item 
  We thank
  Kendall Fortney,
  Tessa Lawler,
  and
  Tabia Tanzin Prama
  for user advice,
\item 
  and Zachary Dalisky
  for software design advice.
\end{textblock}
\begin{textblock}
\item
  This article was assisted in preparation by Samuel F. Rosenblatt in his
  personal capacity.
\item
  The opinions expressed in this article are the author's own and do
  not reflect the view of the Centers for Disease Control and
  Prevention, the Department of Health and Human Services, or the
  United States Government.
\end{textblock}


%

\clearpage



\end{document}